\documentclass[10pt]{article}
\usepackage[utf8]{inputenc}
\usepackage[T1]{fontenc}
\usepackage{amsmath, amssymb, amsfonts, bbm}
\usepackage[a4paper,margin=0.8in]{geometry}
\usepackage{booktabs, multirow}
\usepackage{xspace}
\usepackage{graphicx}
\usepackage[square,numbers]{natbib}
\usepackage{authblk}
\usepackage{url}
\usepackage{algorithm}
\usepackage{algpseudocode}
\usepackage{soul}
\usepackage[table]{xcolor}
\usepackage{subcaption}

\usepackage{fancyhdr}
\definecolor{cvprblue}{rgb}{0.21,0.49,0.74}
\usepackage[pagebackref,breaklinks,colorlinks,allcolors=cvprblue]{hyperref}

\newcommand{\tempo}{TEMPO\xspace}
\newcommand{\unet}{U\mbox{-}Net\xspace}

\title{TEMPO: Global Temporal Building Density and Height Estimation from Satellite Imagery}

\author[1]{Tammy Glazer\thanks{Equal contribution.}}
\author[1]{Gilles Q. Hacheme$^*$}
\author[1]{Akram Zaytar$^*$}
\author[1]{Luana Marotti}
\author[1]{Amy Michaels}
\author[1]{Girmaw Abebe Tadesse}
\author[1]{Kevin White}
\author[1]{Rahul Dodhia}
\author[2]{Andrew Zolli}
\author[1]{Inbal Becker-Reshef}
\author[1]{Juan M. Lavista Ferres}
\author[1]{Caleb Robinson}

\affil[1]{Microsoft AI for Good Research Lab, Redmond, WA, USA}
\affil[2]{Planet Labs PBC, San Francisco, CA, USA}

\date{}
\begin{document}

\makeatletter
\let\oldtable\table
\let\endoldtable\endtable
\let\oldtabular\tabular
\let\endoldtabular\endtabular
\makeatother

\maketitle

\begin{abstract}
We present \tempo, a global, temporally resolved dataset of building density and height derived from high-resolution satellite imagery using deep learning models. We pair building footprint and height data from existing datasets with quarterly PlanetScope basemap satellite images to train a multi-task deep learning model that predicts building density and building height at a 37.6-meter per pixel resolution. We apply this model to global PlanetScope basemaps from Q1 2018 through Q2 2025 to create global, temporal maps of building density and height. We validate these maps by comparing against existing building footprint datasets. Our estimates achieve an F1 score between 85\% and 88\% on different hand-labeled subsets, and are temporally stable, with a 0.96 five-year trend-consistency score. TEMPO captures quarterly changes in built settlements at a fraction of the computational cost of comparable approaches, unlocking large-scale monitoring of development patterns and climate impacts essential for global resilience and adaptation efforts.
\end{abstract}

\begin{figure*}[thb]
    \centering
    \includegraphics[width=1.0\textwidth]{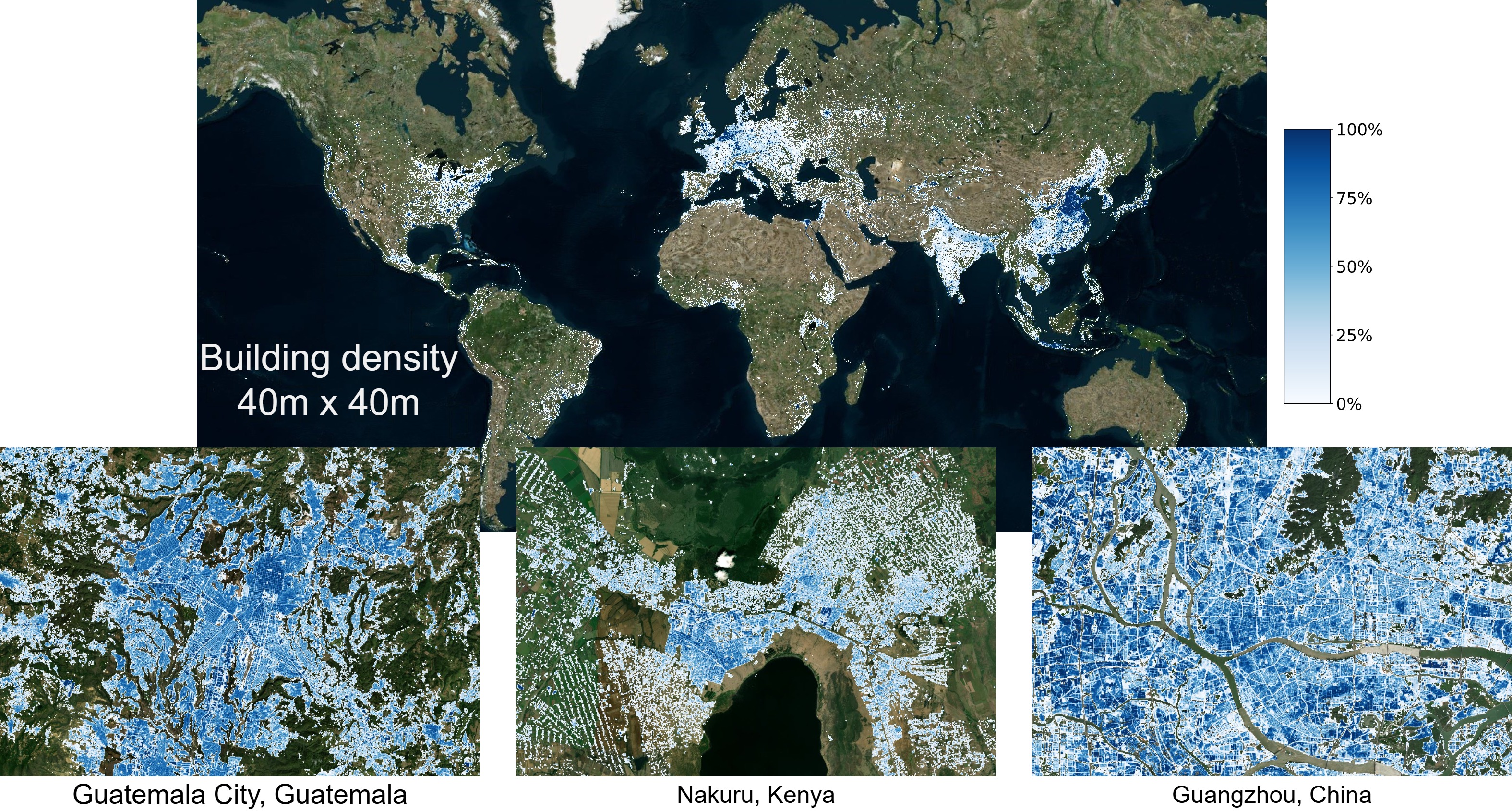}
    \caption{Estimated global building density for Q2 2025.}
    \label{fig:placeholder}
\end{figure*}

\section{Introduction}

Global, temporally consistent building density and height estimates are essential for a wide range of applications, from urban planning and infrastructure development to disaster response and public health~\cite{dobson2000landscan,tatem2017worldpop}. Building data reveals human settlement patterns, informing decisions about transportation, utility networks, and resource distribution~\cite{Oostwegel2025,Sekara2025,Shi2024}. In rapidly growing and environmentally vulnerable regions, accurate maps support disaster resilience and humanitarian efforts~\cite{che20243d, li2024glamour}. Population modeling and demographic analyses rely on signals of building expansion to capture population shifts due to migration, displacement, and urbanization~\cite{boo2022high, zhu2024global}. Monitoring building changes over time is essential for understanding urban growth, informal settlement expansion, and post-disaster recovery, all of which are key factors in sustainable development.

Despite its importance, global building modeling remains challenging due to variations in building characteristics, construction materials, and land use patterns. These challenges are compounded by uneven data availability across regions, where ground truth are disproportionately concentrated in wealthier and urbanized regions leaving vast, resource constrained areas underrepresented. Even state-of-the-art building datasets struggle to achieve consistent performance across visually diverse geographies.~\cite{zhu2024global, wang2018mapping}.

Recent advancements in remote sensing and machine learning have enabled the development of geospatial products that infer building presence from satellite imagery (Table \ref{tab:building_datasets}). However, these datasets often lack the necessary spatial resolution, spatial and temporal and coverage, update cadence, or multi-dimensionality needed to support real-time decision-making. Many existing approaches rely on coarse-resolution data, leading to inaccuracies such as the assignment of buildings to uninhabitable areas or the uniform distribution of buildings across contiguous grid cells~\cite{stevens2015disaggregating, gaughan2013high}. Datasets used to understand urbanization such as OpenStreetMap (OSM) are often incomplete and inconsistent across regions, limiting their utility for large-scale analysis. Further, static representations of built infrastructure lack the temporal dimension needed for real-time monitoring.

In this paper, we introduce \tempo, the first quarterly updated, global building dataset at a 37.6-meter spatial resolution derived from satellite imagery. We train a single multi-task gridded regression model with a shared encoder and two task-specific heads that jointly estimate density and height each quarter from PlanetScope visual basemap inputs using a combination of labeled data from Overture Maps~\cite{OvertureMapsFoundation} and Google Open Buildings 2.5D~\cite{GoogleOpenBuildingsTemporal}. We validate our model's estimates by comparing them to existing large scale building products and hand labeled building footprint data, and compute temporal consistency metrics for our predictions over time.

We release \tempo's global Q4 2023 building density and height estimates at 100-meter resolution plus quarterly series for Nakuru (KE), Guangdong (CN), Guatemala Dept (GT), Bamako (ML), and Lusaka (ZM) from Q2 2020 to Q2 2025 at \url{https://github.com/microsoft/buildings}.

\begin{figure}[bth]
    \centering
    \includegraphics[width=0.7\linewidth]{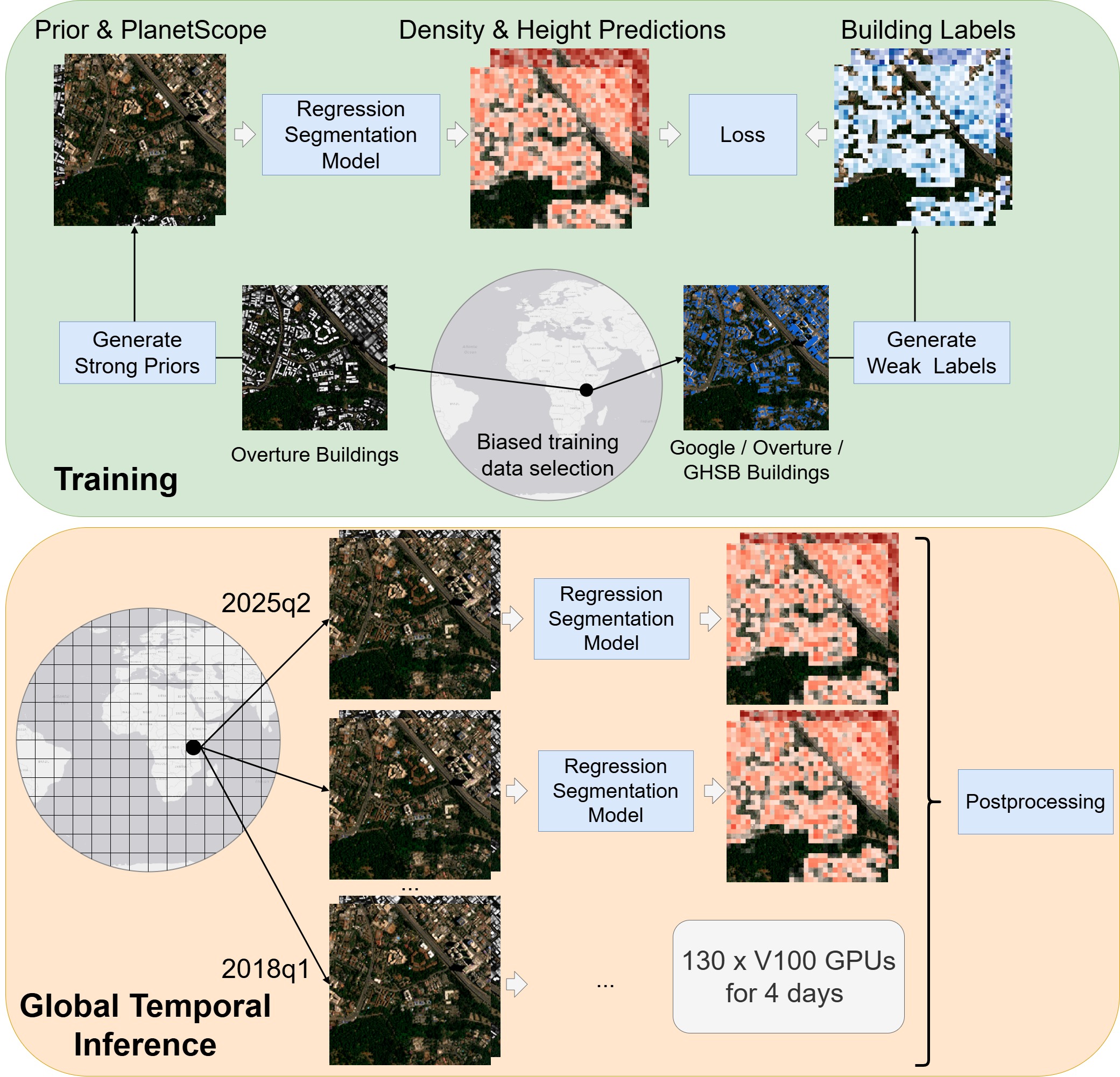}
    \caption{Our methods are split into a \textit{training workflow} (\textbf{top}) and \textit{inference workflow} (\textbf{bottom}). The \textit{training workflow} consists of fitting a multi-task building density \& height segmentation model from weak labels sampled around the globe. The \textit{inference workflow} consists of running the model on nearly 1PB of quarterly PlanetScope basemap imagery from Q1 2018 through Q2 2025 with post-processing to improve temporal consistency.}
    \label{fig:overview}
\end{figure}

\section{Data}

\subsection{Planet Imagery} \label{subsec:data_planet}
We use quarterly PlanetScope visual basemaps from Planet Labs~\cite{planetscope2024} as the primary model input in this work. Planet generates its basemaps by mosaicking imagery captured by the PlanetScope satellite constellation\footnote{Note: RapidEye imagery is used in global basemaps generated between 2019 and 2020.}, which images the entire land surface of the Earth every day~\cite{planetscope2024}. The highest quality scenes taken over an area of interest during a three-month period are combined into a single layer, optimizing for appearance. The final product is a global set of color-corrected, orthorectified RGB imagery at a 4.7-meter per pixel spatial resolution\footnote{While PlanetScope imagery is captured at a 3-meter per pixel resolution, PlanetScope basemaps have a spatial resolution of 4.7-meters per pixel (i.e. the size of a pixel at zoom level 15 at the equator in a slippy tile map).}. These data are well-suited to monitoring infrastructure changes over time and space as they minimize the effects of cloud cover, haze, and topographic variation. The basemaps are distributed as a uniform grid of 4096 x 4096 pixel \textit{quads} covering the entire Earth.

Planet also distributes a \textit{usable data mask} (UDM) with each \textit{quad} of imagery. A UDM classifies each pixel based on image clarity. Images captured prior to Q3 2019 do not have accompanying UDMs. For images captured between 2019 and 2021, UDMs are 8-bit images that encode pixel clarity. Pixels that contain clouds, or are missing, suspect, or damaged, are considered unusable~\cite{planetspecs2023}. For images captured after January 2022, UDMs are multi-band GeoTIFFs, where bands map to pixel classifications: clear, cloud, haze, cloud shadow, or snow~\cite{planetUDM21}. These classifications, along with model confidence scores, are generated by a supervised segmentation model that runs on 4-band top-of-atmosphere radiance assets.

While PlanetScope visual basemaps are designed to provide consistent, high-resolution global imagery, they come with limitations for analytical use. Blending multiple images over a three-month period can mask real-time changes, resulting in temporal latency. The process of mosaiking multiple scenes can also produce visible blending artifacts, such as mismatched textures or boundaries. Variations in sensor calibrations, viewing angles, and illumination conditions between composited scenes can produce geometric and radiometric inconsistencies. Moreover, it is relatively common for scenes to contain haze, shadows, or clouds that obscure the underlying landscape, particularly in tropical, coastal, and mountainous areas. For example, according to data derived from Planet's Q3 2023 UDMs, 16\% of global basemap quads contain clouds, 0.43\% of total pixels are cloudy, and 1.2\% of quads have >10\% cloud coverage.

Finally, Planet deploys periodic updates to the processing algorithms they use to improve image quality, which complicates change detection. Shifts in compositing algorithms and cloud-masking strategies, for instance, affect image brightness and color consistency. In 2022, Planet introduced image sharpening and super-resolution to its Basemap products. These updates, while beneficial for visual analysis, introduce input distribution shifts that must be addressed in modeling.

\subsection{Building Data} \label{subsec:data_buildings}

\begin{table*}[th]
    \centering
    \resizebox{\linewidth}{!}{%
    \begin{oldtabular}{lcccc}
        \toprule
        \textbf{Dataset} & \begin{tabular}[c]{c}\textbf{Spatial}\\ \textbf{Coverage}\end{tabular} & \begin{tabular}[c]{c}\textbf{Spatial}\\ \textbf{Resolution}\end{tabular} & \begin{tabular}[c]{c}\textbf{Temporal}\\ \textbf{Coverage}\end{tabular} & \begin{tabular}[c]{c}\textbf{Temporal}\\ \textbf{Resolution}\end{tabular} \\
        \midrule
        \multicolumn{5}{l}{\textit{Footprint / settlement extent / density products}}\\
        \midrule
        Global Human Settlement Layer (GHSL-BUILT-S) \cite*{pesaresi2024advances} & Global & 10m & 2018 & Static \\
        Global Urban Footprint (GUF) \cite*{esch2017breaking} & Global & 12m & 2011 & Static \\
        GlobalBuildingMap (GBM) \cite*{zhu2024global} & Global & 3m & 2018--2019 & Static \\
        Google High Resolution Buildings \cite*{sirko2021continental} & \begin{tabular}[c]{@{}c@{}}Africa, South/Southeast Asia,\\ Latin America, Caribbean\end{tabular} & 0.5m & Unknown & Dynamic \\
        High-Resolution Settlement Layer (HRSL) \cite*{tiecke2017mapping} & Global & 30m & 2015 & Static \\
        Microsoft Building Footprints \cite*{MicrosoftBuildingFootprints} & Global (except China) & 0.3--0.6m & 2014--ongoing & Dynamic \\
        OpenStreetMap (OSM) & Global (varies) & Varies & Unknown & Dynamic \\
        Overture Maps & Global (varies by region) & Varies & Unknown & Dynamic \\
        SpaceNet7 \cite*{van2021multi} & 59 locations & 3--4m & 2018--2020 & Monthly \\
        World Settlement Footprint (WSF 2019) \cite*{marconcini2021understanding} & Global & 10m & 2019 & Static \\
        \midrule
        \multicolumn{5}{l}{\textit{Height / 2.5D products}}\\
        \midrule
        Google 2.5D \cite*{sirko2023high} & \begin{tabular}[c]{@{}c@{}}Africa, South/Southeast Asia,\\ Latin America, Caribbean\end{tabular} & \begin{tabular}[c]{@{}c@{}}Building-level\\ (footprints at 0.5m)\end{tabular} & 2016--2023 & Yearly \\
        World Settlement Footprint 3D (WSF3D) \cite*{esch2022world} & Global & 90m & 2011--2013 & Static \\
        GHSL Height (GHSL-H, 2018) & Global & 10m & 2018 & Static \\
        GlobalBuildingAtlas (GBA.Height) \cite*{zhu2025globalbuildingatlas} & Global & \begin{tabular}[c]{@{}c@{}}Building-level\\ Building-level\end{tabular} & 2019 & Static \\
        3D-GloBFP \cite*{che20243d} & Global & 0.3--0.6m & 2020 & Static \\
        \midrule
        \textbf{\tempo building density and height} & Global & 40m & Q2 2020--Q2 2025 & Quarterly \\
        \bottomrule
    \end{oldtabular}
    }
    \caption{Comparison of global building footprint/density and height datasets.}
    \label{tab:building_datasets}
\end{table*}

Existing datasets (Table \ref{tab:building_datasets}) aim to map buildings across the globe at a variety of spatial and temporal resolutions from different input data sources and geolocation techniques. Most of these datasets are static and cover a specific point in time -- for example, the Global Urban Footprint dataset is created from Synthetic Aperture Radar (SAR) data collected in 2011. Others, such as Microsoft Building Footprints and Google High Resolution Buildings, are dynamically updated (e.g. when new high-resolution optical imagery is available), but do not provide timestamps for the addition of individual buildings. Community-driven open mapping projects such as OpenStreetMap, though precise in certain locations, have inconsistent completeness by region and are subject to participation bias.

Notably, Open Buildings 2.5D~\cite{GoogleOpenBuildingsTemporal} (Google 2.5D) is the only dataset available with both broad spatial and temporal coverage. Released in 2024, it is produced from yearly time-series of Sentinel-2 imagery and covers Africa, South and Southeast Asia, South America, and the Caribbean. Google 2.5D provides yearly building presence and height estimates from 2016 to 2023, thereby enabling temporal analysis of the built environment. Our work aims to extend this -- providing global building density and height estimates, quarterly.

To estimate building density and height over time, we use multiple building footprint and height datasets in complementary roles -- Overture as a prior, Google 2.5D as weak labels, and others as alignment layers or validation targets -- all processed with a single standardized pipeline. First, we construct a quad index that maps each Planet Level-15 Basemap quad to intersecting tiles or footprint partitions from the reference source. Then, for each quad, we crop to the quad footprint, merge overlaps where needed, reproject to Web Mercator (EPSG:3857), resample to the target $512\times512$ output grid, and normalize band values to $[0,1]$. The resulting products---one band for density or two bands for density and height---are saved as quad-named Cloud-Optimized GeoTIFFs (COGs) on Azure Blob Storage. This uniform grid, file layout, and naming scheme makes all sources interoperable for input priors, weak supervision, alignment evaluation, and validation. We use data from the following sources:

The \textbf{Open Buildings 2.5D (Google 2.5D)}~\cite{sirko2023high} product estimates annual building presence, fractional building counts, and building height estimates over most of Africa, South and Southeast Asia, Latin America, and the Caribbean annually from 2016 to 2023. The dataset was created using a teacher-student deep learning framework, where the teacher learns from static high-resolution satellite data (0.5-meter resolution) and the student uses temporal Sentinel-2 imagery (10-meter resolution) as input, with the teacher's output as the target. The student model, trained on a time series of Sentinel-2 imagery centered around June 30th each year, outputs predictions at 0.5-meter resolution. We use a processed version of Google 2.5D as weak labels in our training pipeline.

The \textbf{Overture Maps Foundation Building (Overture)}~\cite{OvertureMapsFoundation} dataset combines community-contributed data from OpenStreetMap and Esri Community Maps with machine learning-derived data from Google Open Buildings, Microsoft Building Footprints, and recently, Buildings in East Asian Countries. The dataset is constructed using a hierarchical approach that prioritizes community-contributed data first, which is often manually verified, followed by machine learning-derived datasets. Globally, the combined dataset contains 2.3 billion footprints and is the most comprehensive open-source set of building footprint data. We process version v2024-10-23.0 and use this as a model prior (i.e., an additional input band).

The \textbf{Global Human Settlement Layer Built-Surface (GHSL-Built-S)}~\cite{JRC102420} product estimates the proportion of built-up areas at a 10-meter spatial resolution for historical maps, and at 100-meter resolution for projections, from satellite imagery and ancillary sources. Unlike building footprint datasets, which delineate individual structures, GHSL-Built-S estimates the density of built surfaces. We process GHSL-Built-S for alignment analysis.

The \textbf{Global Human Settlement Layer Building Height (GHS-BUILT-H)}~\cite{ghs_built_h_2018} provides a global 100-meter grid of average building heights for the 2018 epoch. Heights are inferred by filtering global DEMs (AW3D30, SRTM30) and updating with shadow markers from a 2018 Sentinel-2 composite using linear regression. The release reports two layers: \emph{Average Net Building Height} (ANBH, meters) and \emph{Average Gross Building Height} (AGBH, \(m^3/m^2\)). We process AGBH for alignment analysis.

The \textbf{World Settlement Footprint 2019 (WSF)}~\cite{esch2022world} dataset provides a global 10-meter resolution binary settlement mask for the year 2019 derived from multitemporal Sentinel\mbox{-}1 (SAR) and Sentinel\mbox{-}2 (optical) imagery. The product exploits temporal statistics of S1- and S2-based indices under the assumption that settlement signals are more temporally stable than other land-cover classes, yielding a consistent delineation of built-up extent that is useful for masking non-settlement areas and benchmarking building-presence estimates. We process WSF for alignment analysis.

\textbf{GlobalBuildingAtlas (GBA)}~\cite{zhu2025globalbuildingatlas} is an open, static, and global dataset that provides individual-building polygons (GBA.Polygon) and per-building height estimates (GBA.Height). GBA.Polygon fuses quality-filtered footprints from existing open sources with machine-derived polygons to produce coverage for more than 2.7\,billion buildings worldwide, while GBA.Height learns heights from PlanetScope time series at 4.7-meter resolution; together these layers enable global models (approximately 2.68\,billion buildings) suitable for large-scale analyses. We process GBA for alignment analysis.

\begin{figure*}[h]
    \centering
    \includegraphics[width=1\textwidth]{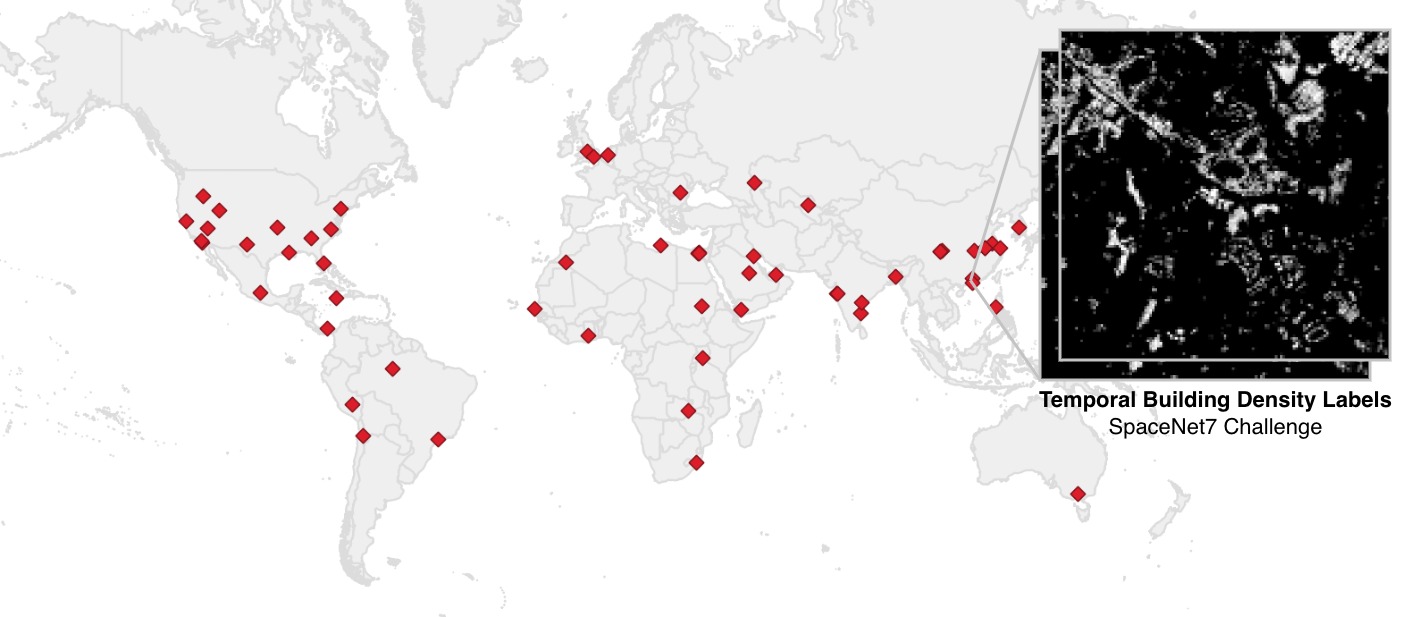}
    \caption{SpaceNet7 building density labels by geographic location.}
    \label{fig:spacenet7}
\end{figure*}

The \textbf{SpaceNet7 Multi-Temporal Urban Development Challenge (SpaceNet7)}~\cite{van2021multi} dataset provides a monthly time-series of high-resolution PlanetScope satellite images hand-annotated with building footprints from 60 locations (see Figure \ref{fig:spacenet7}) and is designed to track trends in rapidly urbanizing areas. SpaceNet7 enables the detection of construction events over time, making it particularly valuable for evaluating urban growth and validating temporal building density models. The labeling process utilized 4.7-meter visual basemaps and higher-resolution images for difficult scenes. We use SpaceNet7 to validate our model predictions.

\section{Methods} \label{sec:methods}
We model global, quarterly building density and height from PlanetScope RGB basemaps. We define \textit{building density} as the percentage of an area of interest covered by human-made structures that are permanently or semi-permanently in one place, following OpenStreetMap's definition of buildings~\cite{OSM_Building_Key}. In addition, we produce a per-pixel \textit{building height} map, where each pixel value is the average building height within that pixel in meters. Both targets are predicted over $8 \times 8$ blocks of basemap pixels, yielding $\approx 37.6 \times 37.6$\ meter spatial resolution at a 3-month (quarterly) temporal cadence. Height values are normalized to $[0,1]$, where $0$ corresponds to $0$\, meters and $1$ to $100$\, meters (extreme values are clipped).

Our modeling approach involves: (i) generating weak labels from publicly available datasets; (ii) training a deep regression model that predicts both density and height maps; (iii) post-processing with auxiliary data sources; and (iv) temporal-consistency smoothing. Each component is described below.

\subsection{Generating Building Labels}

To create quad-level building density and height labels, we establish a one-to-many mapping between Planet \textit{quads} and intersecting Google 2.5D tiles. For each Planet quad, we: (i) reproject the corresponding set of Google 2.5D tiles to the Web Mercator (EPSG:3857) coordinate system using bilinear resampling; (ii) crop and merge overlapping maps to the Planet quad boundary; (iii) downsample the merged files to a standardized $512 \times 512$ pixel grid using average resampling; (iv) save the resulting building density and height estimates as a COG GeoTIFF file. The density values range from 0 (no buildings) to 1 (fully occupied), height values range from $0$ ($0$\, meters) to $1$ ($100$\, meters) with a \textit{no data} value of $-1$ assigned to areas with missing or invalid values. This explicit handling of missing data enables downstream model training to appropriately mask these regions during loss computation.

\subsection{Multi-task Modeling of Buildings} \label{subsec:modeling}

We frame building estimation as two coupled dense-regression tasks--(i) building presence (density) (\%) and (ii) building height (m)--sharing a \unet and using two task-specific convolution heads.

\paragraph{Model Architecture}
We use a modified \unet segmentation architecture~\cite{ronneberger2015} with an ImageNet-pretrained EfficientNet-B6 backbone~\cite{tan2019efficientnet} to estimate building density and height. Specifically, we apply an $8 \times 8$ average pooling operation (with $8 \times 8$ stride and no padding) after the last layer of the \unet decoder, followed by two 1$\times$1 convolution layers. The average pooling step aggregates per-pixel predictions into target estimates, while the 1$\times$1 convolution layers refine the predictions by learning local adjustments based on the aggregated values (e.g. mapping the aggregated values to the $\left[0, 1\right]$ range).

The model takes a single PlanetScope RGB basemap imagery at a 4.7-meter resolution normalized to $[0,1]$ (3 bands) concatenated with Overture Maps building density resampled to the target resolution (1 band) as 4-channel input and produces a two-channel output, representing the estimated coarse density and height for each $8 \times 8$ block of pixels in the input. The Overture Maps ``prior'' provides high-confidence building presence indicators for locations in which Overture data is present.

\paragraph{Sampling Strategy}
Our imagery inputs and building labels are organized according to Planet's quad definitions: $4096 \times 4096$ pixel images paired with $512 \times 512$ pixel building density and height labels. During mini-batch selection we sample quads proportionally to their summed building density labels plus a small constant, ensuring that quads with more buildings are selected more frequently, but that empty quads are also considered. Once a mini-batch of quads is selected, we randomly sample image patches of size $512 \times 512$ paired with corresponding label patches of size $64 \times 64$ within each quad.

\paragraph{Loss Function}
We use a regression-based loss function with a hard sigmoid activation, designed to handle the noise in our weak labels. Since labels may be inaccurate, a standard $MSE$ loss would be overly sensitive to noisy annotations. Instead, we use Huber loss~\cite{huber1992robust}, which treats small errors quadratically and large errors linearly. The loss is given as:

\begin{equation}
L(y, \hat{y}) =
\begin{cases}
\frac{1}{2} (y - \hat{y})^2, & \text{if } |y - \hat{y}| \leq \delta \\
\delta (|y - \hat{y}| - \frac{1}{2} \delta), & \text{otherwise}
\end{cases}
\end{equation}

where $\delta$ is a threshold parameter that controls the transition between the squared loss and the absolute loss, which we set to $0.7$. We apply the loss to the hard sigmoid activation of the predicted density:

\begin{equation}
\sigma_h(x) =
\begin{cases}
0, & \text{if } x < -3, \\
\frac{x+3}{6}, & \text{if } -3 \le x \le 3, \\
1, & \text{if } x > 3.
\end{cases}
\end{equation}

The hard sigmoid activation function enables the model to predict pixel values $0$ and $1$ whereas a standard sigmoid only asymptotically approaches these values. This formulation transforms the problem into a bounded regression task, ensuring that predicted values remain within the valid range of $[0,1]$.

\paragraph{Training}
We train a single \unet, shared across all quarters and years, to predict per\mbox{-}pixel building density and height. Mini\mbox{-}batches are formed by sampling (image, label) patch pairs across years $2020$–$2023$ and quarters (Q1–Q4), so each batch mixes seasons and years. This cross\mbox{-}temporal batching serves as an implicit augmentation, exposing the model to variation in locations, times, and PlanetScope sensor and processing differences, and encourages learning temporally agnostic features rather than quarter\mbox{-}specific cues. We also apply simple geometric augmentation, independently flipping each input patch horizontally and vertically with probability $p=0.5$.

Additionally, we apply spatial dropout (random erasing) on the Overture prior channel to discourage over-reliance. For each training sample, with probability \(p=0.5\) we erase a single rectangular window in the prior band, filling it with zeros while leaving the RGB channels unchanged. The erased window's area fraction \(s\) is drawn uniformly from \([0.2,\,1.0]\) of the patch area and its aspect ratio \(r\) is drawn uniformly from \([0.3,\,3.3]\); the window is placed at a random location and the transform is applied independently per sample. Unlike random pixel-wise dropout, this window masking approach removes priors in spatially coherent regions, simulating realistic scenarios where building footprint data from the Overture prior may be missing in certain geographies.

The model is trained using the ``AdamW'' optimizer with a tuned learning rate of $1.4\mathrm{e}{-3}$ and a batch size of $64$. We apply a cosine annealing learning rate scheduler with a period of $200$ over a total of $1200$ training epochs. Training is performed on approximately $2\%$ of the total available global quad imagery, with $1\%$ of it randomly held out as a validation set. We use a spatial blocking approach~\cite{roberts2017cross} by randomly selecting quads, rather than pixels, for training, validation, and testing. Model training requires about 72 A100-GPU hours to complete. For inference, we use the checkpoint with the lowest validation loss on the third plateau of the cosine schedule.

\subsection{Post-Processing}
\label{subsec:postprocessing}

\begin{figure}[th]
    \centering
    \includegraphics[width=0.8\linewidth]{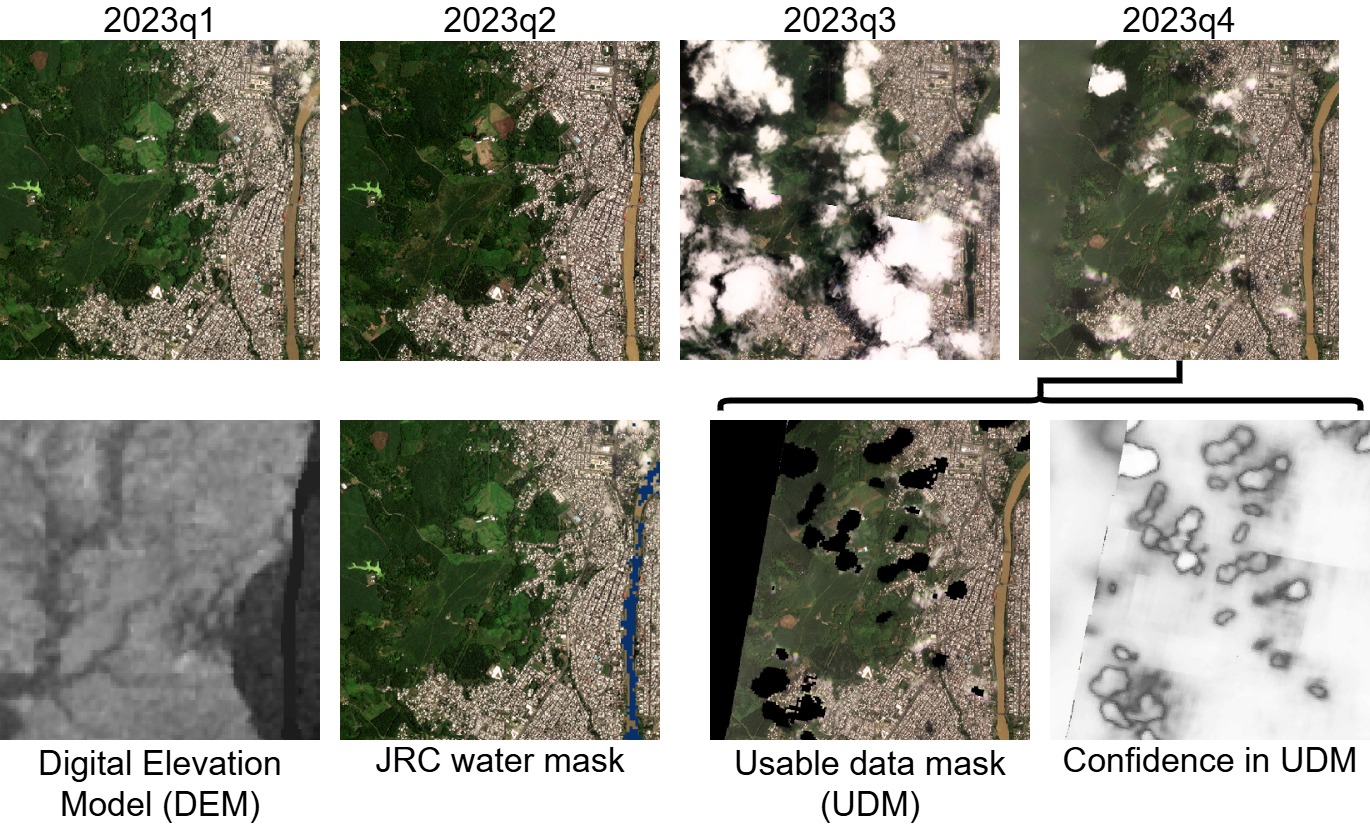}
    \caption{Examples of Planet quarterly imagery for 2023, digital elevation model data, water masks, UDMs, and UDM confidence values over part of Quevedo, Ecuador.}
    \label{fig:ensemble_data}
\end{figure}

To improve the accuracy and temporal stability of building density and height estimates, we apply a post-processing pipeline that involves temporal smoothing and masking of uninhabitable locations to the quarterly model outputs. Figure \ref{fig:ensemble_data} shows the different data layers that are used in our post-processing method.

\subsubsection{Rolling Time-Window Aggregation}

Within each scene, we implement rolling, pixel-level time-window aggregation (Algorithm \ref{alg:rolling_ensemble}). For each pixel in each quarter of predictions, the predictions from the current quarter $t$ are combined with those from previous quarters $[t-1, t-2, t-3]$ and Planet UDM data to create a single smoothed layer. This approach reduces the effects of noise, missing data, and abrupt image variations.

First, we compute a clarity score from 1 to 4 for each pixel in a layer using the UDM and UDM confidence layers. A pixel is counted as high-confidence with a value of $>95\%$. High confidence clear pixels are assigned the highest score (4), followed by low confidence clear pixels (3), low confidence unclear pixels (2), and high confidence unclear pixels (1). The resulting mask of clarity scores is then downsampled to $512 \times 512$ pixels with average resampling.

Next, for each pixel, we apply a rolling time-window aggregation (Algorithm \ref{alg:rolling_ensemble})\footnote{Density values $>2/255$ are considered to represent a valid building.} over a time-series of four predictions to determine the most likely underlying building density estimate for the current timestamp. The same algorithm is applied to building height estimates at the pixel level. All post-processing hyperparameters, including confidence threshold, pixel clarity score, averaging method, and minimum elevation, density, and height values,  were tuned using an automated, multi-trial search framework.

\begin{algorithm}[th]
\caption{Rolling time-window aggregation algorithm.}
\label{alg:rolling_ensemble}
\begin{algorithmic}[1]
\For{each pixel $p$ across the time-series $\{P_t\}_{t=1}^4$}
    \State Use pixel-level agreement to produce $\hat{P}$:
    \If{$\geq 3$ timestamps indicate a building}
        \State Set pixel to the \textbf{median} value of positive pixels.
    \ElsIf{$\geq 3$ timestamps indicate no building}
        \State Set pixel value to $0$.
    \Else
        \If{clarity score $>= 3.5$}
            \State Set pixel value to the \textbf{max} value.
        \ElsIf{clarity score $< 3.5$}
            \State Set pixel value to $0$.
        \Else
            \State Set pixel value to the \textbf{median} value across all 4 timestamps.
        \EndIf
    \EndIf
\EndFor
\end{algorithmic}
\end{algorithm}

\subsubsection{Masking Uninhabitable Locations}

Following the rolling time-window aggregation, we identify locations were buildings are unlikely to exist and use this information to mask false positives. We use the Joint Research Centre's Global Surface Water (GSW) dataset to identify locations likely to contain water~\cite{pekel2016}. Specifically, we use water pixel classifications from the GSW Transitions map, which captures changes in water over time\footnote{GSW maps the distribution of surface water from 1984 to 2021 and provides statistics on the extent and change of water surfaces using classification algorithms on top of Landsat 5, 7, and 8 imagery.}. Any positive building estimate derived from a location known to contain permanent or seasonal water is removed\footnote{JRC GSW Transition classifications of permanent water (1), new permanent water (2), seasonal water (4), new seasonal water (5), seasonal to permanent water (7), and permanent to seasonal water (8) are applied.}.

We also mask predictions in high elevations that are unlikely to contain buildings. As less than $0.004\%$ of the population is known to live above 5,000 meters~\cite{tremblay2021} and the highest known permanently inhabited town in the world sits at 5,100 meters~\cite{west2002highest}, we remove positive building density and height estimates derived from locations above 5,100 meters. We use the NASADEM Global Digital Elevation Model, which provides global topographic data at 1 arc-second horizontal resolution, to identify these locations~\cite{nasadem2020}.

\subsubsection{Ensuring Agreement}

Building height and density values are adjusted to ensure logical consistency between the two layers. Pixels with nonzero building density values are assigned a minimum height of 2.4 meters, consistent with international building code standards for habitable rooms \cite{IBC2021_1208.2}. Height estimates below this threshold typically represent noise, vegetation, or small sheds rather than true structures; enforcing this minimum suppresses spurious low-height signals. Pixels with zero density are assigned zero height, and pixels with nonzero height are assigned a minimum density of 2/255 to ensure that even minimal structures reflect a physically plausible built fraction rather than a numerical artifact.

\subsection{Global Scaling}
\label{subsec:global-scaling}

To operationalize the trained model at planetary scale, we deploy two complementary large-scale compute pipelines: one for distributed inference and another for post-processing. Distributed inference generates a raw global prediction archive (\tempo) from Q2 2018 to Q2 2025, while the post-processing pipeline consumes this archive—together with auxiliary layers such as Planet UDMs, elevation, and water masks—to produce the final, temporally smoothed dataset (\tempo-PP) available from Q2 2020 to Q2 2025~\footnote{The smoothed dataset (\tempo-PP) is only available from Q2 2020, as UDMs were introduced starting Q2 2019.}.

\subsubsection{Distributed Inference}

Once trained, we scale inference globally across all quarterly PlanetScope basemaps from Q1 2018 through Q2 2025 using distributed Azure Machine Learning (AML) jobs. Each run launches a Message Passing Interface (MPI) pool on a V100 cluster; across runs we use 130 NVIDIA V100 GPUs in parallel. A manifest of 916{,}400 PlanetScope quads is evenly sharded by node rank so that each GPU processes a disjoint subset. For each quad, we stream RGB and Overture imagery directly from Azure Blob Storage, normalize bands using the checkpoint's scalers, and run the frozen model with mixed precision. Predictions are written as tiled, compressed GeoTIFFs (float32) in Web Mercator (EPSG:3857), one file per quad with the relevant bands (density and height), and uploaded to Azure Blob Storage. Per-node JSON logs capture allocations, successes, and failures to support safe restarts and targeted re-runs. This procedure yields a raw global prediction layer per quarter; each full-resolution timestamp is approximately $43$TB. We use the 2018--2025 archive for alignment experiments, but only quarters that can be post-processed with UDMs contribute to the \tempo-PP product.

\subsubsection{Parallel Post-Processing}

Post-processing of the global prediction archive is executed using a hybrid parallel and asynchronous framework designed for large-scale geospatial I/O. Rolling time-window aggregation combines predictions from the current quarter $t$ with those from $t-1$, $t-2$, and $t-3$ and relies on Planet UDMs to weight pixel clarity. Because UDMs are only available from Q2 2019 onward, the earliest quarter for which all four windows have UDM support is Q2 2020. Consequently, we run the post-processing pipeline from Q2 2020 through Q2 2025, and \tempo-PP is defined over this range.

Individual tiled GeoTIFFs are read directly from Azure Blob Storage and processed in parallel using a lightweight thread-based executor for concurrent, I/O-bound operations. An asynchronous task scheduler is used to offload blocking read operations (e.g., imagery and UDM retrieval) while we execute lightweight transformations and metadata extraction concurrently. Computationally intensive routines, including temporal differencing and rolling-window aggregation, are distributed across worker threads to maximize CPU utilization. Frequently accessed datasets such as elevation models and water masks are cached in memory and reused across tasks, eliminating redundant reads. This multi-layered concurrency model enables high-throughput post-processing, reducing wall-clock runtime and ensuring scalability over the Q2 2020 -- Q2 2025 archive.

\subsection{Evaluation}
\label{subsec:temporal_quality}

We evaluate \tempo's predictions using both static and temporal criteria. First, we measure the accuracy of per-quarter density and height estimates using regression and classification metrics tailored to the built-environment domain. We then assess temporal consistency to determine how reliably the model captures real-world growth patterns over time.

\subsubsection{Evaluating Static Predictions}

To evaluate regression behavior, we use mean absolute error (MAE), which measures the average absolute deviation between predicted and reference values. Because most pixels globally have zero built fraction, we compute MAE only over locations with strictly positive reference values to avoid artificially lowering the error and to focus on regions containing buildings. To evaluate classification behavior, we threshold continuous density and height outputs and report F1 alongside precision and recall. For height, we compute macro-F1 over discrete height bins so that low-, mid-, and high-rise structures contribute equally. F1 is well-suited for comparing against reference datasets that are available only at specific timestamps or at coarser spatial resolutions, and is robust in the presence of strong class imbalance. We report accuracy and $R^2$ as complementary indicators of calibration and variance explained, but rely on F1 and MAE as our primary evaluation metrics.

\subsubsection{Evaluating Temporal Consistency}

High–quality temporal building maps must be (i) stable in places where the built environment does not change and (ii) sensitive in places that experience real growth. We therefore evaluate stability on annual snapshots and report temporal consistency on density. Each quad prediction map is summarized using non-overlapping tiles of \(k\times k\) pixels with \(k\in\{3,7,10\}\). Tiles that are zero in all years are ignored. We compare Google 2.5D, TEMPO before post-processing, and TEMPO following post-processing (TEMPO-PP) on annual snapshots, using TEMPO's Q4 predictions each year to match Google 2.5D availability.

\paragraph{Temporal Correlation (Year-to-Year)}
For each adjacent year pair available in both TEMPO and Google 2.5D (2020--2021, 2021--2022, 2022--2023), we compute the Pearson correlation across tiles between the two years. Tiles that are zero in both years of a pair are excluded from that pair's correlation. We median-aggregate pair correlation values. This metric helps us measure the stability of yearly layers: if there are no significant false positives or false negatives between the two layers, the Pearson correlation should be close to $1$.

\paragraph{Monotonicity AUC (Tolerant Trend Consistency)}
This metric is based on the assumption that, over the time window of interest, the vast majority of the world is either temporally stable (no building reduction or new construction) or exhibits monotonic change--predominantly increasing building volume, and only rarely net reduction. Temporal patterns that show a large drop followed by a large rise (or vice versa) are therefore likely to reflect temporal inconsistencies or errors rather than real-world dynamics. Hence, layers should be temporally stable, meaning that building volume in most regions should be roughly stable--only increasing or only decreasing. For each tile, we compute year-to-year differences. A tile is called \emph{monotonic within tolerance}~$\tau$ if all year-to-year steps satisfy either (non-decreasing within $\tau$): $\Delta_t \ge -\tau$ for all $t$, or (non-increasing within $\tau$): $\Delta_t \le \tau$ for all $t$, where $\Delta_t$ denotes the change from year $t-1$ to year $t$. We sweep $\tau$ over 100 evenly spaced values in $[0,0.01]$, compute the percentage of tiles that are monotonic at each $\tau$, and then take the area under this monotonicity curve using the trapezoidal rule. A larger AUC indicates that a greater fraction of tiles follow a consistent stable, upward or downward trend even under stricter tolerances, and thus that the layer is more temporally consistent.

\paragraph{Difference Standard Deviation}
We pool all year-to-year differences from all tiles and compute their overall standard deviation. Lower values indicate smoother, less noisy changes over time. This represents a global spread of first differences.

\section{Results}

We evaluate \tempo along four axes: (i) ablations that isolate the contribution of each architectural and training choice; (ii) alignment with external datasets (2018–2025) for both density and height; (iii) building-presence detection on SpaceNet7 across regions and years; and (iv) temporal stability of Q4 annual density estimates. Unless noted otherwise, we report the unmasked Full model (see Section ~\ref{subsec:ablations}), which contains all evaluated modeling components, as the main configuration.

\subsection{Ablations}
\label{subsec:ablations}

\begin{table*}[th]
\centering

\newcommand{\used}{\cellcolor{green!25}}
\newcommand{\notused}{\cellcolor{red!25}}
\newcommand{\dependent}{\cellcolor{white}}

\begin{tabular}{l c c c c c c c c c c}
\toprule
\textbf{Model} & \rotatebox{90}{\textbf{Backbone}} & \rotatebox{90}{\textbf{Weighted}} & \rotatebox{90}{\textbf{Overture}} & \rotatebox{90}{\textbf{Aug}} & \rotatebox{90}{\textbf{Hard Sig}} & \rotatebox{90}{\textbf{2 Heads}} & \rotatebox{90}{\textbf{RGB}} & \rotatebox{90}{\textbf{Agree}} & \multicolumn{2}{c}{\textbf{Loss}} \\
\cmidrule(lr){10-11}
 &  &  &  &  &  &  &  &  & \textbf{Value} & \textbf{$\Delta$ (\%)} \\
\midrule
Full              & \used B6 & \used & \used & \used & \used & \used & \used & \notused & 2.19e-4 & --- \\
\midrule
-1                & \notused B5 & \used & \used & \used & \used & \used & \used & \notused & 2.29e-4 & +5\% \\
-2                & \used B6 & \notused & \used & \used & \used & \used & \used & \notused & 2.68e-4 & +22\% \\
-3                & \used B6 & \used & \notused & \dependent & \used & \used & \used & \notused & 7.58e-4 & \textbf{+246\%} \\
-4                & \used B6 & \used & \used & \notused & \used & \used & \used & \notused & 2.21e-4 & +1\% \\
-5                & \used B6 & \used & \used & \used & \notused & \used & \used & \notused & 2.20e-4 & 0\% \\
-6                & \used B6 & \used & \used & \used & \used & \notused & \used & \notused & 3.19e-4 & \textbf{+46\%} \\
\midrule
Mask disagreements & \used B6 & \used & \used & \used & \used & \used & \used & \used & 2.46e-4 & +12\% \\
Minimal data      & \used B6 & \used & \notused & \dependent & \used & \used & \used & \used & 7.80e-4 & \textbf{+256\%} \\
Light model       & \notused B4 & \used & \used & \notused & \used & \used & \used & \notused & 2.64e-4 & +21\% \\
Only Overture     & \used B6 & \used & \used & \notused & \used & \used & \notused & \notused & 6.14e-4 & \textbf{+180\%} \\
\bottomrule
\end{tabular}

\caption{Ablation study evaluating individual component contributions. \colorbox{green!25}{Green} indicates enabled components, \colorbox{red!25}{Red} indicates disabled components, \fcolorbox{black}{white}{White} indicates components automatically disabled due to dependencies. Components include: \textbf{Backbone}: ImageNet-pretrained EfficientNet variants (B4/B5/B6) used as U-Net encoders. \textbf{Weighted}: Training samples weighted by summed building-density vs. uniform (random) sampling. \textbf{RGB}: Indicates the use of PlanetScope 4.7-meter RGB bands as input. \textbf{Overture}: Overture density prior stacked as additional input channel. \textbf{Aug}: Black-box inpainting applied to the Overture band to encourage learning from the RGB (requires Overture to be enabled). \textbf{Hard Sig}: Indicates the use of the Hard sigmoid activation function in final layer (instead of sigmoid function). \textbf{Two Heads}: Dual output heads for density and height prediction. \textbf{Agree}: Whether we restrict training to pixels where Google 2.5D and Overture labels agree, with pixels that disagree masked during loss computation. The \textbf{Full model} uses all components except \textbf{Agree}. Rows -1 through -6 ablate single components from \textbf{Full}. We report Huber Loss after the first cosine cycle. $\Delta$ (\%) shows relative degradation compared to the \textbf{Full} model. Lower loss is better.}
\label{tab:ablations}
\end{table*}

We run a controlled ablation study starting from a \textbf{Full} configuration that uses an EfficientNet-B6 \textbf{Encoder}, density-\textbf{weighted} quad sampling, 4-channel input (PlanetScope \textbf{RGB} plus \textbf{Overture} prior), spatial dropout \textbf{augmentation} on the Overture band, a \textbf{hard-sigmoid}-bounded Huber loss, and dual density/height \textbf{heads}. Rows $-1$ to $-6$ each disable exactly one component of this recipe: swapping the backbone to EfficientNet-B5 ($-1$), switching to uniform (random) sampling ($-2$), removing the Overture prior band ($-3$), disabling Overture augmentations while keeping the prior ($-4$), replacing hard sigmoid with the sigmoid activation function ($-5$), and using a single shared head instead of dual heads ($-6$). Additional rows test combinations: for instance, \textbf{Mask disagreements} only trains on pixels where Google 2.5D and Overture agree; \textbf{Minimal data} trains on agreements without the Overture prior band; \textbf{Light model} replaces the EfficientNet-B6 encoder with EfficientNet-B4 and omits Overture augmentation; and \textbf{Only Overture} removes RGB so the model relies solely on the Overture prior. All variants reuse the same labels, optimizer, and training schedule as the \textbf{Full} model.

Table~\ref{tab:ablations} reports validation loss at the first plateau of our cosine schedule on the Q4 2023 validation set, which in practice preserves the same ordering as full training and therefore serves as a reliable proxy for final performance. The Full configuration achieves the lowest loss, and every ablation either degrades performance or has negligible effect. The largest degradations come from removing the Overture prior or relying solely on it without RGB, confirming that PlanetScope RGB and Overture provide complementary signals: RGB captures temporal visual context while Overture supplies a strong, spatially aligned density prior where imagery is noisy or small buildings are hard to resolve. Decoupling density and height into dual heads also brings a substantial gain over a shared head, indicating that the two regression tasks benefit from shared features but require separate output parameterizations. Density-weighted sampling consistently improves over uniform (random) sampling, emphasizing the need for training over built-up regions. In contrast, hard-sigmoid vs.\ sigmoid and Overture augmentations show minimal impact on validation loss, but we retain them for their benefits on downstream detection behavior and robustness in regions with unreliable priors. Finally, combined ablations highlight that data quality matters more than architecture: aggressively restricting supervision (minimal data) or dropping RGB harms performance far more than using a lighter backbone, showing that high-quality, complementary labels and inputs are the main drivers of accuracy.

\subsection{Alignment with External Datasets}

\begin{table*}[th]
    \centering
    \small

    \begin{subtable}[t]{\textwidth}
        \centering
        \caption{Detection (F1 at 1\% threshold)}
        \begin{tabular}{llcccccccc}
            \toprule
            & & \multicolumn{5}{c}{Density (reference datasets)} & \multicolumn{3}{c}{Height (reference datasets)} \\
            \cmidrule(lr){3-7} \cmidrule(lr){8-10}
            Model & Year & Google & Overture & GHSL & GBA & WSF & Google & GHSL & GBA \\
            \midrule
            \multirow{6}{*}{TEMPO} & 2018 & 0.852 & --    & --    & --    & --    & 0.601 & 0.392 & --    \\
             & 2019 & 0.857 & --    & --    & --    & 0.742 & 0.599 & --    & 0.482 \\
             & 2020 & 0.858 & --    & 0.672 & --    & --    & 0.602 & --    & --    \\
             & 2021 & 0.861 & --    & --    & --    & --    & 0.598 & --    & --    \\
             & 2022 & 0.862 & --    & --    & --    & --    & 0.600 & --    & --    \\
             & 2023 & 0.859 & 0.780 & --    & 0.532 & --    & 0.599 & --    & --    \\
            \midrule
            \multirow{4}{*}{TEMPO-PP} & 2020 & \textbf{0.872} & --    & \textbf{0.690} & --    & 0.757 & \textbf{0.746} & 0.528 & 0.663 \\
             & 2021 & \textbf{0.872} & --    & --    & --    & --    & \textbf{0.740} & --    & --    \\
             & 2022 & \textbf{0.873} & --    & --    & --    & --    & \textbf{0.753} & --    & --    \\
             & 2023 & \textbf{0.870} & \textbf{0.800} & --    & \textbf{0.550} & --    & \textbf{0.756} & --    & --    \\
            \bottomrule
        \end{tabular}
    \end{subtable}

    \vspace{0.75em}

    \begin{subtable}[t]{\textwidth}
        \centering
        \caption{Regression (MAE)}
        \begin{tabular}{llcccccccc}
            \toprule
            & & \multicolumn{5}{c}{Density (\%)} & \multicolumn{3}{c}{Height (m)} \\
            \cmidrule(lr){3-7} \cmidrule(lr){8-10}
            Model & Year & Google & Overture & GHSL & GBA & WSF & Google & GHSL & GBA \\
            \midrule
            \multirow{6}{*}{TEMPO} & 2018 & 1.40 & --    & --    & --    & --    & 0.11 & 0.03 & --   \\
             & 2019 & 1.39 & --    & --    & --    & 14.26 & 0.11 & --   & 0.14 \\
             & 2020 & 1.40 & --    & 2.21 & --    & --    & 0.11 & --   & --   \\
             & 2021 & 1.42 & --    & --    & --    & --    & 0.12 & --   & --   \\
             & 2022 & 1.45 & --    & --    & --    & --    & 0.11 & --   & --   \\
             & 2023 & 1.46 & 3.71 & --    & 0.56 & --    & 0.11 & --   & --   \\
            \midrule
            \multirow{4}{*}{TEMPO-PP} & 2020 & \textbf{1.37} & --    & 2.27 & --    & 14.26 & 0.45 & 0.28 & 1.21 \\
             & 2021 & \textbf{1.41} & --    & --    & --    & --    & 0.44 & --   & --   \\
             & 2022 & \textbf{1.44} & --    & --    & --    & --    & 0.45 & --   & --   \\
             & 2023 & \textbf{1.45} & \textbf{3.55} & --    & 0.61 & --    & 0.44 & --   & --   \\
            \bottomrule
        \end{tabular}
    \end{subtable}

\caption{Alignment of TEMPO and TEMPO-PP building density and height predictions with external reference datasets, 2018--2023 (quarterly metrics averaged). Upper table: detection (F1 at 1\% threshold); lower table: regression (MAE in \% for density, m for height). For reference datasets predating 2020, TEMPO-PP metrics use the nearest available year (2020). \textbf{Bold values} indicate TEMPO-PP improvements over TEMPO for the same year and metric. ``--'' indicates unavailable reference data for that year. Reference timing: Google Open Buildings 2.5D uses the corresponding year; Overture Maps uses the 2023 snapshot; GHSL uses 2020 (density) and 2018 (height) epochs; WSF is 2019; GBA uses 2023 (density) and 2019 (height).}
\label{tab:alignment}
\end{table*}

Table~\ref{tab:alignment} evaluates cross-source alignment between \tempo, \tempo-PP, and external building products. We treat the reference datasets as labels, compute quarterly metrics, and report yearly averages for 2018--2023 under two setups: classification and regression. For classification, we threshold \tempo density at $0.01$ and threshold targets at $0$ (Overture) or $0.01$ (all other references) to report detection F1; for height, we compute a macro-F1 by detecting three bins---$(10^{-4}, 3\,\mathrm{m})$, $(3, 10\,\mathrm{m})$, and $(10, 30\,\mathrm{m})$---and averaging. For regression, we report MAE over pixels where the reference is strictly $> 0$ to avoid zero-inflation. Each reference is evaluated at its native epoch (Google 2.5D by year; Overture 2023; GHSL density 2020 and height 2018; WSF 2019; GBA density 2023 and height 2019), and we show results for both raw predictions (\tempo) and the post-processed outputs (\tempo-PP) under the same protocol to quantify the effect of temporal smoothing.

Overall, \tempo-PP achieves the highest detection alignment on the datasets used by the model---Google 2.5D (weak labels) and Overture (prior). Across independent references, detection alignment remains strong, particularly against GHSL and WSF, indicating broad cross-dataset similarity. \tempo-PP is also stable over time, with no quarter-specific collapse; this is most evident against the temporally varying Google 2.5D baseline, where both detection and MAE remain steady. For regression, the errors are best interpreted as calibration gaps: MAE is generally low across references except for WSF, where higher MAE likely reflects scale differences rather than missed detections. Rolling time-window aggregation delivers the clearest benefit for detection, substantially improving cross-dataset alignment by suppressing false positives in uninhabited regions via masking. In contrast, post-processing has little impact on density regression and slightly degrades height regression, consistent with density-guided adjustments that yield a different height product. In summary, the post-processed product (\tempo-PP) achieves the strongest cross-dataset detection alignment, while density aligns most with Google 2.5D, Overture, and WSF, and height aligns most with Google 2.5D and GBA.

\subsection{Building Presence Detection}

\begin{table}[th]
\centering
\begin{tabular}{@{}ccccccc@{}}
\toprule
\textbf{Set}                  & \textbf{Model} & \textbf{Precision} & \textbf{Recall} & \textbf{F1} & \textbf{Accuracy} & \textbf{R2} \\ \midrule
\multirow{5}{*}{Global South} & TEMPO          & 0.79               & 0.92            & 0.85        & 0.74              & 0.71        \\
                     & TEMPO-PP & 0.90 & 0.86 & \underline{\textbf{0.88}} & \underline{\textbf{0.78}} & 0.71 \\
                     & GHSL             & 0.66 & \textbf{0.98} & 0.79 & 0.65 & 0.00 \\
                     & Overture         & \textbf{0.96} & 0.52 & 0.67 & 0.51 & 0.31 \\
                     & Google 2.5D      & 0.77 & 0.93 & 0.84 & 0.73 & \underline{\textbf{0.72}} \\ \midrule
\multirow{4}{*}{All} & TEMPO            & 0.73 & 0.92 & 0.81 & 0.68 & 0.70 \\
                     & TEMPO-PP & 0.86 & 0.84 & \underline{\textbf{0.85}} & \underline{\textbf{0.74}} & \underline{\textbf{0.70}} \\
                     & GHSL             & 0.62 & \textbf{0.96} & 0.76 & 0.61 & 0.00 \\
                     & Overture         & \textbf{0.94} & 0.44 & 0.60 & 0.43 & 0.25 \\ \bottomrule
\end{tabular}%
\caption{Performance of models on SpaceNet7 derived building density labels over two subsets: \textit{All} contains the latest observation across time from the full dataset (59 ``quads''), and \textit{Global South} is the latest observation per quad restricted to tiles for which Google 2.5D provides predictions (predominantly Global South). Precision, Recall, F1, and Accuracy are computed after binarizing building density at \(>0\) (any built vs. none), while \(R^2\) is computed on the continuous building-density values. Higher is better for all metrics. \textbf{Boldface} marks the best value in each column; underlining highlights balanced metrics (F1, Accuracy, \(R^2\)).}

\label{tab:spacenet7_results}
\end{table}

We evaluate building-presence detection against SpaceNet7–derived density labels on two subsets: \textit{Global South} (quads within Google 2.5D's coverage) and \textit{All} (latest observation per quad across all 59 sites). For each site, we map SpaceNet7 masks to their Planet quads, crop to the quad footprint, reproject, and resample to the model's output grid. Classification metrics are computed after binarizing density values: $>0.01$ for TEMPO and Google 2.5D, and $>0$ for GHSL, Overture, and SpaceNet-7. We report Precision, Recall, F1, Accuracy, and $R^2$. Results are shown for the Global South domain (top block) and globally (bottom block); see Table~\ref{tab:spacenet7_results}.

The patterns are consistent. Overture yields the highest precision, reflecting its high-confidence footprints, while GHSL achieves the highest recall, consistent with its propensity to over-extend. In the Global South, our post-processed product (TEMPO-PP) attains the best F1 and Accuracy by a wide margin, with Google 2.5D slightly ahead on $R^2$. Globally, TEMPO-PP widens its lead, delivering the top F1, Accuracy, and $R^2$. These gains stem from our training pipeline, heuristics-based masking, and rolling time-window aggregation, which suppress false positives and stabilize noisy quarters. In short, TEMPO-PP matches or surpasses contemporaneous products on building detection while offering quarterly cadence and global coverage.

\subsection{Temporal Stability}

\begin{table*}[th]
\centering
\small
\setlength{\tabcolsep}{6pt}
\begin{tabular}{@{}l ccc ccc ccc@{}}
\toprule
& \multicolumn{3}{c}{\textbf{Corr} ($k$)} 
& \multicolumn{3}{c}{\textbf{AUC} ($k$)} 
& \multicolumn{3}{c}{\textbf{STD} ($k$)} \\
\cmidrule(lr){2-4}\cmidrule(lr){5-7}\cmidrule(lr){8-10}
\textbf{Dataset} & 3 & 7 & 10 & 3 & 7 & 10 & 3 & 7 & 10 \\
\midrule
Google 2.5D    & .989 & .995 & .996 & .701 & .851 & .896 & .014 & .007 & .005 \\
TEMPO      & .991 & .996 & .997 & .753 & .904 & .936 & .011 & .005 & .004 \\
TEMPO-PP  & \textbf{.995} & \textbf{.998} & \textbf{.998} &
             \textbf{.833} & \textbf{.943} & \textbf{.963} &
             \textbf{.009} & \textbf{.004} & \textbf{.003} \\
\bottomrule
\end{tabular}
\caption{Temporal stability metrics at window sizes $k \in \{3,7,10\}$. Higher Corr and AUC are better; lower STD is better.}
\label{tab:temporal_stability}
\end{table*}

For temporal stability, we compare raw and smoothed \tempo density against the only temporally varying external reference, Google 2.5D. Because Google 2.5D is annual, we use \tempo Q4 predictions each year and restrict evaluation to the spatial intersection covered by Google 2.5D. Windows serve as our data points: we fix a window size \(k\in\{3,7,10\}\), tile each quad with non-overlapping \(k\times k\) windows, and, within each window, average the valid density pixels. Repeating this for each year yields, for every window, a 1D window-signal that traces the evolution of mean density over 2020--2023 (windows that are zero in all years are ignored). We then compute stability on these window-signals for each dataset separately: (i) Pearson correlation across adjacent years, (ii) AUC of tolerant monotonicity, and (iii) the standard deviation of year-to-year differences (lower is better).

As window size increases, noise is reduced and consistency metrics increase across the board. Most metrics are near saturation, with monotonicity AUC remaining the most discriminative; even so, a clear advantage emerges. Raw \tempo exceeds Google 2.5D on all three stability measures, and post-processing further improves it. In summary, the post-processed temporal estimates achieve the strongest stability and consistency over 2020--2023 relative to the Google 2.5D baseline.

\begin{figure}[th]
    \centering
    \includegraphics[width=0.9\linewidth]{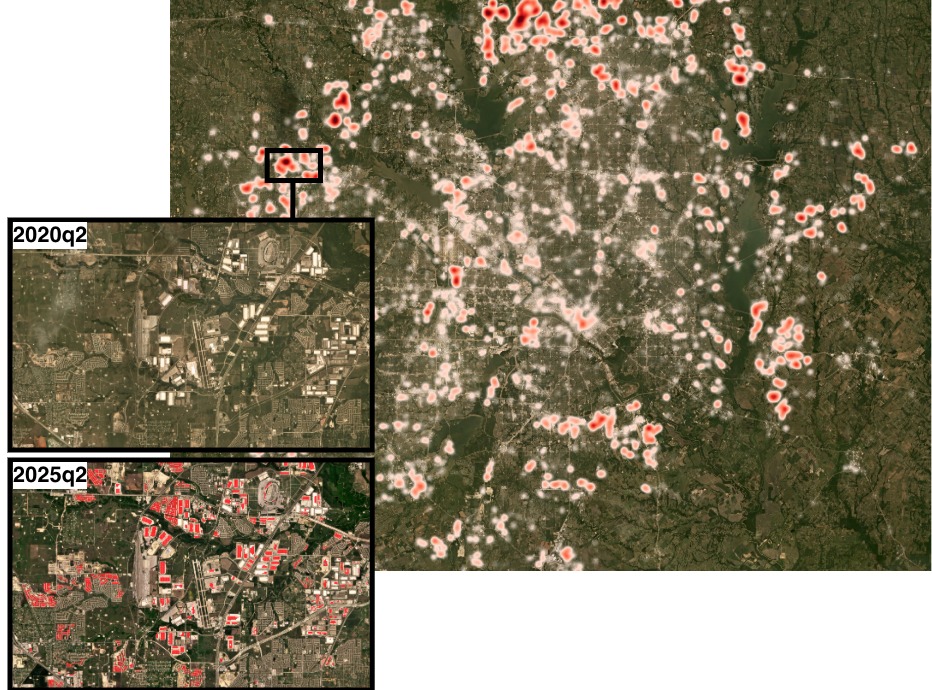}
    \caption{Locations in the greater Dallas, Texas, area that experienced in the top 5\% of urban growth from Q2 2020 to Q2 2025. We highlight a concentration of new warehouse and logistics infrastructure north of Forth Worth, Texas.}
    \label{fig:dallas}
\end{figure}

\subsection{Change Detection Case Studies}

We evaluate TEMPO's utility in identifying change over a 5-year period, from Q2 2020 to Q2 2025, in two spatially distinct regions: Dallas, Texas, USA, and southeast Chad. For each area of interest (AOI), we read in building density and height estimates from each timestep, then for each temporal image pair we then a per-pixel volume proxy and the delta in volume between the later and earlier timestamps. To identify statistically significant growth, we pool positive growth values across the AOI and set a global threshold at the 95th percentile to generate binary inclusion masks. Finally, we vectorize masks with 8-connectivity to produce change polygons per tile.

\subsubsection{Peri-Urban Expansion in Dallas, Texas, USA (2020-2025)}

TEMPO-derived growth clusters highlight a clear pattern of outward industrial and residential expansion along the Dallas-Fort Worth periphery between Q2 2020 and Q2 2025 (Figure~\ref{fig:dallas}). High-volume change is concentrated in logistics corridors and newly urbanizing zones north and west of the urban core, particularly around major highway interchanges and airport-adjacent industrial parks. These findings align with recent studies documenting acclerated post-pandemic warehouse construction and housing development across North Texas \citep{dallasfed2025industrial,texasdemocenter2025northtexas}. The limited signal of infill growth within the inner city ring, contrasted with extensive peri-urban densification, supports observations that metropolitan Dallas is characterized by horizontal, low-density sprawl rather than vertical redevelopment. TEMPO's volumetric data clearly highlights how recent urban growth has spread outward along the city's edges, mirroring known demographic shifts and land-use patterns.

Zooming in, between Q2 2020 and Q2 2025, TEMPO reveals concentrated new construction along the Alliance-Blue Mound-Elizabeth Creek corridor north of Fort Worth. Figure~\ref{fig:dallas} shows recently completed logistics, warehousing, and e-commerce facilities. Development clusters align with Intermodal Parkway, adjacent to the BNSF Alliance Intermodal Facility and Alliance Airport (AWF). This finding is consistent with the ongoing build-out of the AllianceTexas inland-port district. Additions span firms such as FedEx Ground, Dollar General, ITS Logistics, Amazon, and Harbor Freight, illustrating how industrial land absorption and freight-oriented urbanization have driven peri-urban expansion in North Texas during the post-pandemic period \citep{dallasfed2025industrial,texasdemocenter2025northtexas}.

\subsubsection{Encampment Expansion in Southeastern Chad (2020-2025)}

\begin{figure}[tbh]
    \centering
    \includegraphics[width=0.9\linewidth]{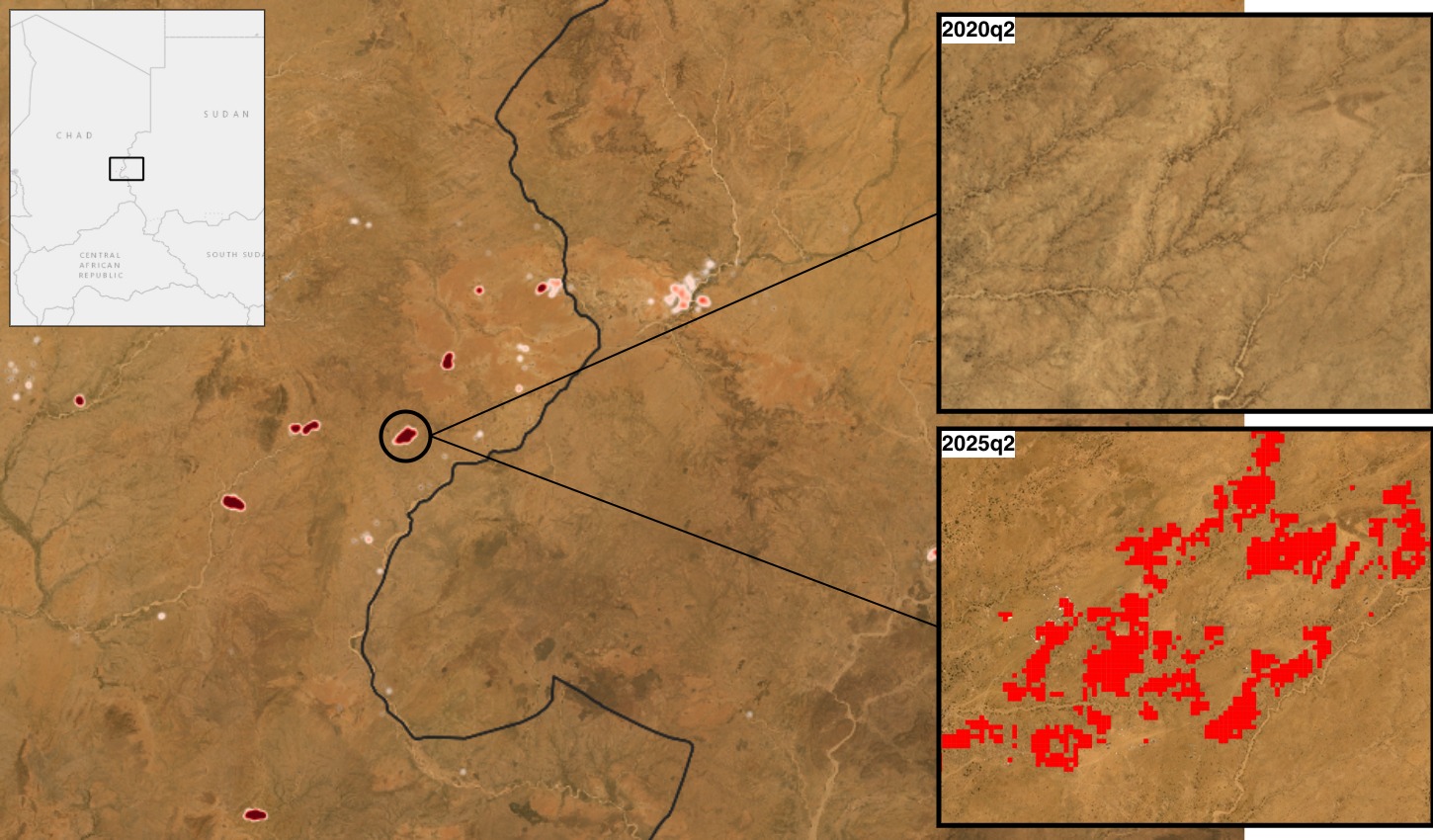}
    \caption{Red areas signify new structures detected between 2020-2025. Camp Métché is a large refugee camp in eastern Chad that houses tens of thousands of people who fled the conflict in Sudan. It was established in late 2023.}
    \label{fig:chad}
\end{figure}

TEMPO identifies pronounced volumetric growth concentrated around Camp Métché, near the Chad-Sudanese border. The change clusters in Figure~\ref{fig:chad} correspond to the rapid physical expansion of refugee and internally displaced person (IDP) settlements visible in satellite imagery, where formerly sparse, semi-arid terrain now exhibits dense, grid-like patterns of built structures and tents. This expansion aligns spatially with known humanitarian encampments reported by the UN Refugee Agency and crisis mapping initiatives during the escalation of the Darfur conflict and subsequent cross-border displacement into Chad between 2023 and 2024. The growth signal reflects the emergence of new shelter blocks, service areas, and access roads within the camp's footprint, as well as sites extending eastward from the main settlement \citep{unhcr2025sudan_chad_update,msf2025metche}.

Together, these patterns demonstrate TEMPO's capacity to capture humanitarian-driven land transformation in non-urban environments, revealing how displacement crises contribute to measurable built-area change in data-scarce regions of the Sahel.

\section{Conclusion and Limitations}

We present \tempo, the first global, quarterly maps of building density and height. \tempo applies a lightweight \unet (EfficientNet-B6) to PlanetScope basemaps guided by Overture priors. Trained on only $2\%$ of the world, the model scales globally via distributed inference and produces temporally consistent results with a simple post-processing step based on rolling time-window aggregation. This lightweight framework enables global inference with a single model and modest computational resources, reducing the cost and complexity of temporal building mapping. Our experiments show that data quality—selective sampling, strong priors, and sensible post-processing—matters more than architectural complexity for cutting-edge building detection. We also outline a practical validation recipe when large, held-out ground truth is unavailable, combining cross-dataset alignment and temporal consistency checks.

\textbf{Limitations.} Weak supervision inherits spatial biases from source footprints; performance varies by region, and local fine-tuning may be required. Single-view 4.7-meter RGB images can miss small, low-contrast, or isolated structures, creating a ``difficulty bias'' that disadvantages traditional housing common in many low-resource communities. Height is less accurate in the tails (very tall buildings), likely due to rarity in training and single-view inference. Temporal noise complicates change analysis, and building declines from damage or partial demolition can be difficult to separate from seasonal or sensor artifacts. Finally, aggregation improves stability but introduces latency: majority voting over four quarters can delay recognition of new construction by up to one quarter.

Most of the datasets we use are public, and the pipeline can be reproduced with Sentinel-2 mosaics at coarser resolution. Looking ahead, we will explore region-adaptive training, predicting tall buildings via hybrid sampling and footprint-aware height masking when resampling, and multi-view inputs over recent quarters to exploit shadows and better detect building reductions. We hope \tempo serves as a simple, scalable training and inference recipe that others can adapt for both global and locally tuned building mapping applications.

{
    \small
    \bibliographystyle{unsrt}
    \bibliography{main}
}

\end{document}